\documentclass[letterpaper, 10 pt, conference]{ieeeconf}
\IEEEoverridecommandlockouts
\usepackage{amsmath,amssymb,amsfonts}
\usepackage{algorithmic}
\usepackage{graphicx}
\usepackage{textcomp}
\usepackage[x11names]{xcolor}
\usepackage{booktabs}
\usepackage{multirow}
\usepackage{subcaption}

\makeatletter
\let\NAT@parse\undefined
\makeatother
\usepackage{hyperref}
\hypersetup{
colorlinks=true, linkcolor=red, citecolor=blue, urlcolor=black}

\usepackage[capitalise]{cleveref} 

\begin{document}
\bstctlcite{IEEEexample:BSTcontrol} 

\title{\LARGE \bf ISSAFE: Improving Semantic Segmentation in Accidents by Fusing Event-based Data}
\author{Jiaming Zhang$^{1}$, Kailun Yang$^{1}$ and Rainer Stiefelhagen$^{1}$
\thanks{This work was supported in part through the AccessibleMaps project by the Federal Ministry of Labor and Social Affairs (BMAS) under the Grant No. 01KM151112, in part by the University of Excellence through the ``KIT Future Fields'' project,  and in part by Hangzhou SurImage Company Ltd.
\textit{(Corresponding author: Kailun Yang.)}}
\thanks{$^{1}$Authors are with Institute for Anthropomatics and Robotics, Karlsruhe Institute of Technology, Germany (e-mail: firstname.lastname@kit.edu).}
\thanks{Code and dataset will be made publicly available at: \url{https://github.com/jamycheung/ISSAFE}}
}
\maketitle

\begin{abstract}
Ensuring the safety of all traffic participants is a prerequisite for bringing intelligent vehicles closer to practical applications. The assistance system should not only achieve high accuracy under normal conditions, but obtain robust perception against extreme situations. However, traffic accidents that involve object collisions, deformations, overturns, etc., yet unseen in most training sets, will largely harm the performance of existing semantic segmentation models. To tackle this issue, we present a rarely addressed task regarding semantic segmentation in accidental scenarios, along with an accident dataset \textit{DADA-seg}. It contains 313 various accident sequences with 40 frames each, of which the time windows are located before and during a traffic accident. Every 11th frame is manually annotated for benchmarking the segmentation performance. Furthermore, we propose a novel event-based multi-modal segmentation architecture \textit{ISSAFE}. Our experiments indicate that event-based data can provide complementary information to stabilize semantic segmentation under adverse conditions by preserving fine-grain motion of fast-moving foreground (crash objects) in accidents. Our approach achieves +8.2\% mIoU performance gain on the proposed evaluation set, exceeding more than 10 state-of-the-art segmentation methods. The proposed \textit{ISSAFE} architecture is demonstrated to be consistently effective for models learned on multiple source databases including Cityscapes, KITTI-360, BDD and ApolloScape. 
\end{abstract}

\section{Introduction}
Intelligent Vehicles (IV) and Advanced Driver Assistance Systems (ADAS) benefit from breakthroughs in deep learning algorithms. In particular, image semantic segmentation can provide pixel-wise understanding of driving scenes, containing object categories, shapes, and locations. In recent years, many state-of-the-art segmentation models~\cite{zhao2017PSPNet, chen2018deeplabv3plus, yuan2019OCRNet, yin2020dnl} have achieved impressive successes in accuracy on major segmentation benchmarks. Other works~\cite{romera2017erfnet, orsic2019swiftnet} centered on improving the efficiency of the model, in order to deploy real-time semantic segmentation on mobile platforms. 

\begin{figure}
    \centering
    \includegraphics[width=0.99\columnwidth]{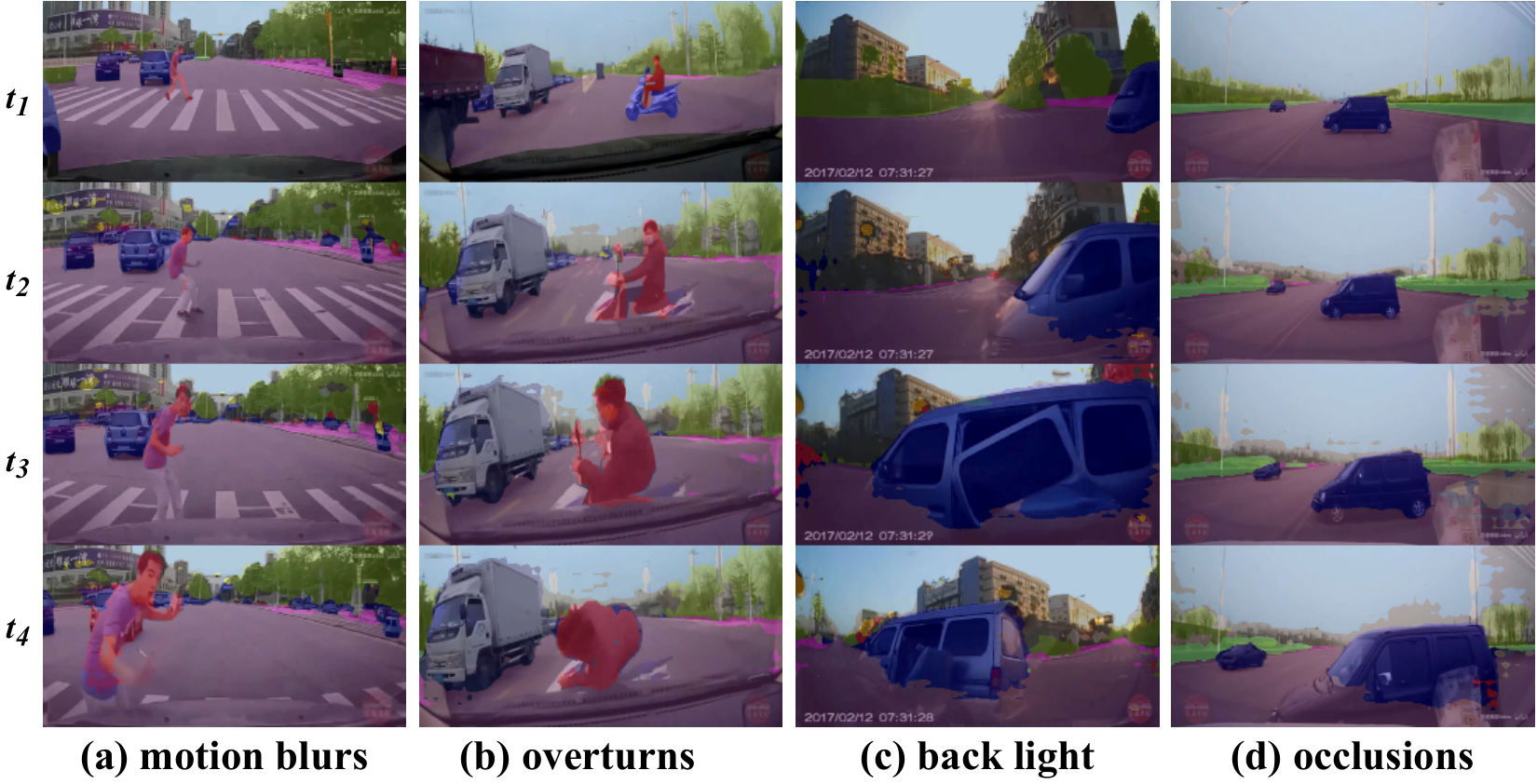}
    \vskip-1ex
    \caption{\small Accident sequences from our \textit{DADA-seg} dataset include diverse hazards (\textit{e.g.} motion blur, overturns, back light, object occlusions). From top to bottom are timestamps before and during an accident, where the $t_{1}$ frame is the ground truth for quantitative evaluation, and the others are predictions of our model.}
    \label{fig:DADA_SegExample}
\vskip -5ex
\end{figure}
Unfortunately, driving environments in the real world are more complicated than most existing datasets, divided into normal, critical and accidental situations. In addition to natural related factors in the normal driving scene, such as weathers and illuminations, many human-centered crisis incidents from other traffic participants may occur. For example, vehicles overtaking irregularly, pedestrians dashing across the road, or cyclists riding out of lanes, these critical situations are all potential causes of traffic accidents, but never seen in vision datasets. Furthermore, the initial accident scene ahead is also defined as an accidental situation, such as an overturned truck or a knocked down motorcycle lying on the road, which should be correctly recognized by passing vehicles in time, only then can pileups be avoided. However, these abnormalities will result in a large and sharp performance drop of the segmentation models when taken from public training imagery to the wild.

To satisfy the rigorous requirements of safety-relevant IV systems, a segmentation model should be thoroughly tested on some edge cases to verify its robustness and reliability. Thus, this paper propose an alternative benchmark based on a new task, namely \textit{Semantic Segmentation in Accidents (SSA)}. Being a supplement to classic benchmarks~\cite{cordts2016cityscapes, zendel2018wilddash}, our evaluation samples are collected from real-world traffic accidental situations which involve highly dynamic scenes and extremely adverse factors. Some cases are shown in \cref{fig:DADA_SegExample}, covering diverse situations: motion blur while the pedestrian is dashing across the road, overturning of the motorcyclist during the collision, back-lighting at the intersection, and occlusions by windshield reflection. As far as known to us, these factors are still challenging for most segmentation algorithms and even harmful to their performance. The objective of creating this benchmark is to provide a set of edge cases (critical and accidental) for testing the robustness of vision models before deployment in real applications.

In addition to traditional cameras, event cameras are bio-inspired novel sensors, such as the Dynamic Vision Sensor (DVS)~\cite{patrick2008DVS_sensor}, that encode \textit{changes} of intensity at each pixel asynchronously and have the characteristics like higher dynamic range ($>120 dB$), high time resolution ($1 MHz$ clock or $\mu s$ timestamp), and are not affected by motion blur~\cite{gallego2019event_survey}. Hence, we consider that event cameras are more sensitive to capture the motion information during driving, especially for fast-moving objects (foreground) in extreme or accident scenarios, where classic cameras delay between frames. In low-lighting environments, event cameras still stably bring sufficient perceptual information. Underlying these assumptions, complementary information can be extracted from the event-based data to address shortcomings of the intensity image in both normal and abnormal scenes.

Finally, as a preliminary exploration on this new task, we propose a light-weight ISSAFE architecture, which can serve as an \textit{Event-aware Fusion (EF)} model to process RGB and event data, or an \textit{Event-aware Domain Adaptation (EDA)} model to bridge the source (normal) and target (accident) datasets. 
In accordance with our ISSAFE architecture, the robustness of SSA can be significantly improved. In summary, our main contributions are:
\begin{itemize}

    \item We present a rarely solved task concerning \textit{Semantic Segmentation in Accidents (SSA)}, with the ultimate goal to robustify the perception algorithm against abnormal situations during highly dynamic driving.
    \item We provide an accompanying accident dataset \textit{DADA-seg}, of which the evaluation set was manually annotated for benchmarking the robustness of SSA.
    \item We propose a multi-modal segmentation architecture \textit{ISSAFE} to exploit complementary features from event-based data according to two approaches, \textit{i.e.} EF and EDA. Comprehensive comparisons and ablation studies are conducted between various models, datasets and data modalities.
\end{itemize}

\section{Related Works}
\subsection{Semantic Segmentation}
Since FCN~\cite{long2015FCN} used fully convolutional layers for pixel-wise prediction on images, a massive number of models~\cite{zhao2017PSPNet, chen2018deeplabv3plus, yuan2019OCRNet} have achieved remarkable performance in image semantic segmentation. In addition to high accuracy, other works, such as ERFNet~\cite{romera2017erfnet} and SwiftNet~\cite{orsic2019swiftnet}, proposed simplified architectures to improve the efficiency. Regarding generalizability, Domain Adaptation (DA) strategies were extensively applied to adapt the segmentation algorithm to new scenes~\cite{luo2019CLAN}. For example, the day-night conversions in~\cite{sun2019seeclear} and the adaptations between diverse weathers like rainy~\cite{pizzati2020i2i_UDA} and snowy~\cite{liu2019SS_snowy_DA} scenes. However, apart from these natural conditions in real driving scenes, there are many uncontrollable factors during the interaction with other traffic participants. The core purpose of our work is to fill the gap of semantic segmentation in abnormal situations.

Any ambiguity in machine vision algorithms may cause fatal consequences in autonomous driving, thus the robustness testing conducted in diverse driving conditions is essential. For this reason, WildDash~\cite{zendel2018wilddash} provided ten different hazards, such as blurs, underexposures or lens distortions, as well as negative test cases against the overreaction of segmentation algorithms. 
However, in order to extend the robustness test from ordinary to accident scenarios, we create an accident dataset DADA-seg. Those critical or accidental scenes are more difficult by having a large variety of adverse hazards. 

On the other hand of improving robustness, some solutions constructed a multi-modal segmentation model by fusing additional information, such as depth in RFNet~\cite{sun2020rfnet}, thermal information in RTFNet~\cite{Y2019RTFNet_rgb_thermal} and optical flow in~\cite{rashed2019optical_flow}. Differing from these classic modalities, in this work, event-based data will be explored as a novel auxiliary modality.

\subsection{Event-based Vision}
Event cameras are increasingly used in visual analysis due to their complementary features to traditional cameras, such as High Dynamic Range (HDR), no motion blur, and response in microseconds~\cite{gallego2019event_survey}. Instead of capturing an image in a fixed rate, event cameras asynchronously encode the intensity change at each pixel with the position, time, and polarity: $(x, y, t, p)$. Typically, for processing in a convolutional network, the original event stream is converted into an image form,
such as a two-channel event frame in~\cite{maqueda2018eventframe_2channel}, a four-dimensional grid in~\cite{zhu2018ev-flownet} and a Discretized Event Volume (DEV) in~\cite{zhu2019unsupervised}.

Based on these image-like representations, Ev-SegNet~\cite{alonso2019Ev-SegNet} was trained on an extended event dataset DDD17~\cite{binas2017ddd17}, whose labels were generated by a pre-trained model and only contain 6 categories. In contrast, our models are trained with the ground-truth labels of Cityscapes in all 19 classes. Additionally, instead of stacking images in the input stage, event data will be adaptively fused with the RGB image through our attention module, which is more effective for combining two heterogeneous modalities.

While labeled event data for semantic segmentation is scarce in the state of the art, other works leveraged the existing labeled data of images by simulating their corresponding event data.
EventGAN~\cite{zhu2019eventgan} presented a self-supervised approach to generate events from associated images using only modern GPUs. In this work, we utilize the EventGAN model to extend the datasets by generating their associated event data, so as to investigate the benefit of event sensing in dynamic accident scenes. Finally, the EDA between both datasets is performed by fusing RGB images and the synthesized events.

\section{Methodology}

\subsection{Task Definition}\label{method:task_definition}
To evaluate the robustness of semantic segmentation models, we create a new task: \textit{Semantic Segmentation in Accidents (SSA)}. Besides, an associated evaluation set is provided for quantitative analysis. All test samples are edge cases collected from real-world traffic accidents and contain adverse situations. We explicitly study the robustness in challenging accident scenarios based on the assumption that the less performance degradation of the algorithm in this unseen dataset, the better its robustness. 

\subsection{Accident Dataset}\label{method:dataset}
\textbf{Dataset Annotation.} Our proposed dataset DADA-seg is selected from the large-scale DADA-2000~\cite{conf/ieeeits/FangYQXWL19_DADA} dataset, which was collected from mainstream video sites. To extend it to the semantic segmentation community, we performed additional laboratory work for processing all 2000 sequences in two stages. In the first stage, sequences with large watermarking or low resolution were removed, while most of the typical adverse scenes were retained, such as those with motion blur, over/underexposures, weak illuminations, occlusions, etc. All other different conditions are described in \cref{tab:condition_distribute}. Concentrating on accident scenes, we remain the 10 frames before the accident and 30 frames during the accident. After selection, the final DADA-seg dataset composes of 313 sequences with a total of 12,520 frames at a resolution of 1584$\times$660. 

\begin{table}[]
\centering
\caption{\small Distribution of total 313 sequences from DADA-seg dataset under conditions in terms of light, weather and occasion.}
\label{tab:condition_distribute}
\begin{tabular}{c|c@{\hskip 3pt}c|c@{\hskip 3pt}c|c@{\hskip 3pt}c@{\hskip 3pt}c@{\hskip 3pt}c@{\hskip 3pt}}
\toprule
\multirow{2}{*}{\textbf{DADA-seg}}   & \multicolumn{2}{c|}{\textbf{Light}} & \multicolumn{2}{c|}{\textbf{Weather}} & \multicolumn{4}{c}{\textbf{Occasion}}     \\ \cmidrule{2-9}
           & day & night & sunny & rainy & highway & urban & rural & tunnel \\ \midrule \midrule
\#sequence & 285 & 28    & 297   & 16    & 32      & 241   & 38    & 2      \\ \bottomrule
\end{tabular}
\vskip -4ex
\end{table}

In the second stage, based on the same 19 classes as defined in Cityscapes, we manually perform full pixel-wise annotation on every 11th frame of 313 sequences by using the polygons to delineate individual semantic classes, as shown in the $t_1$ frame in~\cref{fig:DADA_SegExample}. After labeling, our DADA-seg dataset includes 313 labeled images for the quantitative analysis of SSA and 12,207 unlabeled images for EDA between the normal and abnormal imagery. Comparatively, all images of our dataset are taken in broad regions by different cameras from various viewpoints. Besides, all sequences focus on accident scenarios, composing of normal, critical, and accidental situations. In such a way, the evaluation performed on DADA-seg dataset reflects more thoroughly the robustness of semantic segmentation algorithms.

\subsection{Event-based Data}
\textbf{Event Representation.} Event cameras asynchronously encode an event at each individual pixel ($x$, $y$) at the corresponding triggering timestamp $t$, if the change of logarithmic intensity $L$ in time variance $\Delta t$ is greater than a preset threshold $C$: 

\begin{equation}
    L(x, y, t) - L(x, y, t - \Delta t) \geq  pC, \; p\in \left \{ -1, +1 \right \}
\end{equation}
where polarity $p$ indicates the positive or negative direction of change. A typical volumetric representation of a continuous event stream with size $N$ is a set of 4-tuples: 
\begin{equation}
    V = \left \{ e_i \right \}_{i=1}^N, \textup{where} \; e_i = (x_i, y_i, t_i, p_i). 
\end{equation}

However, it is still arduous to transmit the asynchronous event spike to the convolutional network by retaining a sufficient time resolution. Hence, we perform a dimensionality reduction operation in the time dimension~\cite{zhu2019eventgan}. The original volume is discretized with a fixed length for positive and negative events separately, and each event is locally linearly embedded to the nearest time-series panel. According to the number of positive time bin $B^+$, a discretized spatial-temporal volume $V^+$ is represented as:
\begin{equation}
    \Tilde{t_{i}} =(B^+-1)\left(t_{i}-t_{1}\right) /\left(t_{N}-t_{1}\right),
    \vspace{-0.2cm}
\end{equation}
\begin{equation}
    V^+(x, y, \Tilde{t_{i}}) =\sum_{i}^{B^+} \max \left(0,1-\left|t-\Tilde{t_{i}}\right|\right),
\end{equation}
where the $\Tilde{t_{i}}$ is the embedded timestamp of event $e_i$ according to the time bins $B^+$.
When both positive and negative volumes are concatenated along time dimension, the entire volume is represented as $V \in \mathbb{R} ^{B \times W\times H}, B=B^+ + B^-$, where $B$, $W$ and $H$ are the total number of time bins, the width and height of spatial resolution, respectively. The detailed setting of time bins $B\in\{1, 2, 18\}$ will be discussed in the experiments section.

\begin{figure}
	\centering
    \includegraphics[width=0.85\linewidth]{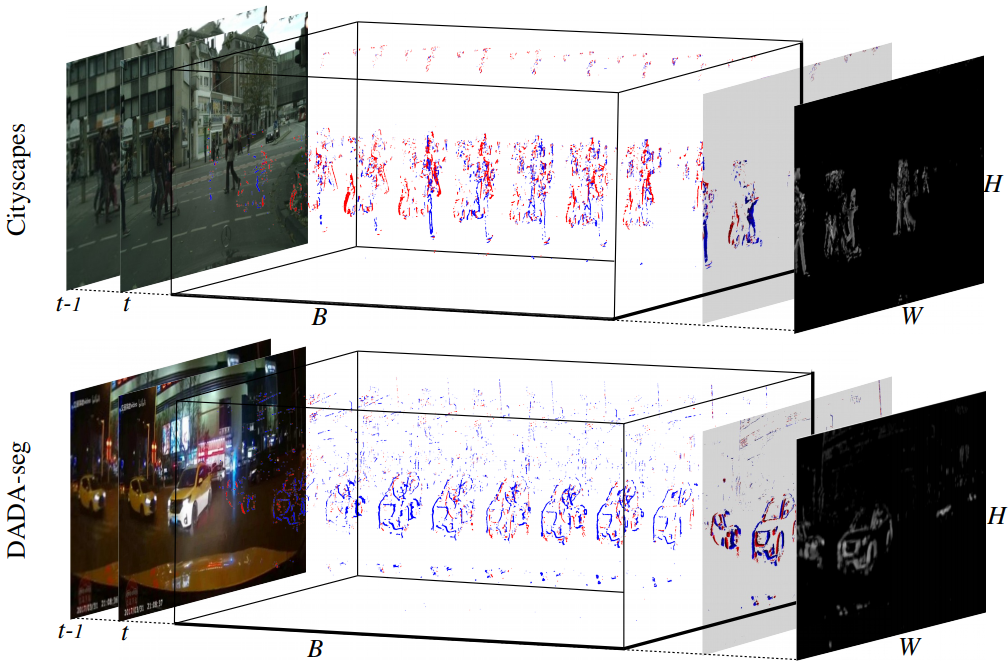}
	\caption[]{\small Visualization of generated event data in $B\times H \times W$ space, which denote the time bins, image height and width. From left to right are RGB image pair ($I_{t-1}$, $I_{t}$), different event representations: event volume, event polarity frame and event grayscale frame, where blue and red colors indicate positive and negative events.} 
	\label{fig:event_gen}
\vskip -4ex
\end{figure}

\textbf{Event Data Synthesis.} Bringing event data to SSA task, there is still a lack of event-based dataset with semantic annotations. Thus, we utilize the identical EventGAN~\cite{zhu2019eventgan} model to synthesize highly reliable event data of all mentioned datasets. For this, two adjacent RGB image frames are required as inputs. Different from the fixed frame rate (17Hz) in Cityscapes~\cite{cordts2016cityscapes}, the sequence in the DADA-seg dataset was acquired with diverse cameras and frame rates, which means that its synthesized event data vary from the intensity of motion due to different time intervals. After verification, the penultimate frame was selected and stacked with its anchor frame for event data synthesis. Two cases of the generated event data are visualized in~\cref{fig:event_gen}.
It can be seen how event data benefits the sensing in driving scenes with moving objects or in low-lighting environments, meanwhile providing higher time resolution in volumetric form.

\begin{figure*}
	\centering
		\begin{subfigure}[b]{0.95\textwidth}   
		\centering 
		\includegraphics[width=\textwidth]{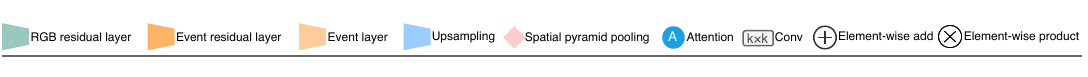}
	\end{subfigure}
	\vskip -1ex
	\begin{subfigure}[b]{0.35\textwidth}   
		\centering 
		\includegraphics[width=\textwidth]{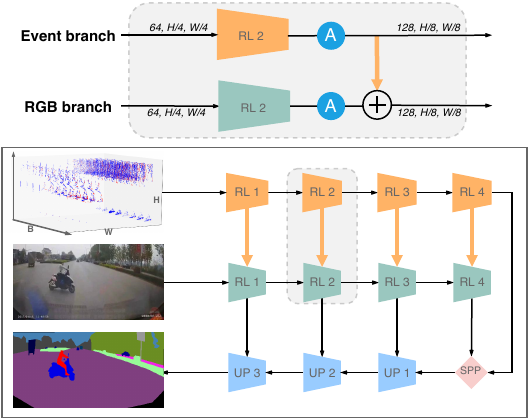}
		\caption[]{\small S2D: event fusion from sparse to dense}    
		\label{fig:S2D}
	\end{subfigure}
	\hskip 0.1\textwidth
	\begin{subfigure}[b]{0.35\textwidth}
		\centering
		\includegraphics[width=\textwidth]{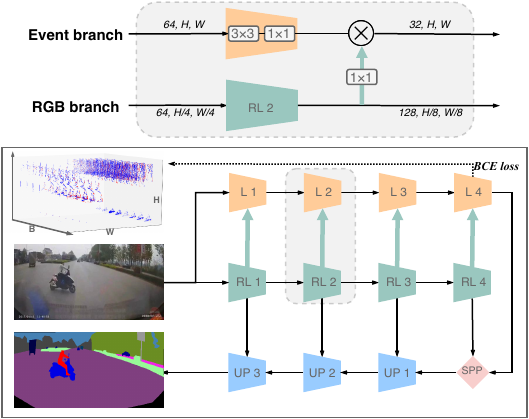}
		\caption[]{\small D2S: event fusion from dense to sparse}    
		\label{fig:D2S}
	\end{subfigure}
	\vskip -1ex
	\caption[]{\small Model architectures of two different event fusion strategies. In (a), event data is fused to RGB branch adaptively from sparse to dense, while in (b) event data is extracted from dense image and learned from the sparse ground truth.} 
	\label{fig:event_fuse}
\vskip -4ex
\end{figure*}

\subsection{ISSAFE: Event-aware Fusion}
According to the characteristics of event data flow from sparse to dense or vice versa, two diverse EF approaches are mainly designed and explored to excavate complementary informative features from the event data. 

\textbf{S2D: Sparse-to-Dense.} An intuitive fusion approach is treating the sparse event data as input and extracting its dense feature, similar to a normal RGB image branch. In this paper, we mainly explore the adaptive fusion of these two different modalities between layers. As shown in~\cref{fig:S2D}, the S2D fusion model includes dual branches, \textit{i.e.} RGB branch and event branch, constructed with the ResNet-18~\cite{he2016resnet} backbone for maintaining a real-time speed. 
Inspired by the design of SwiftNet~\cite{orsic2019swiftnet} and RFNet~\cite{sun2020rfnet}, a channel-wise attention module is employed between layers of both branches, in which the motion features are emphasized in the event branch and added element-wise into the RGB branch. In other words, the higher time resolution from event data complements the motion-related features in the blurred RGB image. Additionally, its HDR enhances the over/underexposure image. 
While the image feature map is termed as $\mathcal{F}_{i} \in \mathbb{R} ^{C \times H\times W}$ and the event feature map as $\mathcal{F}_{e} \in \mathbb{R} ^{C' \times H\times W}$, the S2D fused feature $\mathcal{F}_{S2D} \in \mathbb{R} ^{C \times H\times W}$ by channel-wise attention is represented as:
\begin{equation}
    \mathcal{F}_{S2D}=\mathcal{F}_{i} \otimes \sigma_i\left[f(\mathcal{F}_{i})\right] + \mathcal{F}_{e} \otimes \sigma_e\left[g(\mathcal{F}_{e})\right],
\end{equation}
where both $f(\cdot)$ and $g(\cdot)$ are composed of the adaptive global pooling and $1\times1$ convolution operations, the $\sigma_i(\cdot)$ and $\sigma_e(\cdot)$ denote Sigmoid activate functions for the image and event feature map. 
After four residual layers, the event feature serves as an additional stream in the Spatial Pyramid Pooling (SPP) module~\cite{zhao2017PSPNet} and will be concatenated with other high-level features for long-range context sensing. Finally, a light-weight decoder, composing of 3 upsampling modules with 1$\times$1 skip connections from the RGB branch, will align different levels of features for the final prediction. 

\textbf{D2S: Dense-to-Sparse.} On the other hand, inspired by the video restoration from a single blurred image and the event data like \cite{pan2019video_from_event, jin2018video_from_blurred_image}, we alternatively leverage the D2S fusion approach, as shown in~\cref{fig:D2S}. Varying from the classic residual layer in the previous S2D fusion mode, a more light-weight encoder with 4 layers is constructed as the event branch. 
Each layer only contains a 3$\times$3 and a 1$\times$1 convolutional kernel, which is more effective to extract features from dense to sparse and capable of processing at a higher spatial resolution. After the initial convolution of ResNet-18~\cite{he2016resnet}, while the RGB branch encodes higher-level features at smaller resolutions with \{4, 8, 16, 32\} downsampling rates and \{64, 128, 256, 512\} channels, the event branch deactivates the non-event features according to the higher-level semantic features from the RGB branch. Meanwhile, the event branch gradually shallows event channels in the order of \{64, 32, 16, 8\} for final event prediction, which also enables event processing at the full resolution. 
The dense RGB feature map $\mathcal{F}_{i} \in \mathbb{R} ^{C \times H\times W}$ and the sparse event feature map $\mathcal{F}_{e} \in \mathbb{R} ^{C' \times H\times W}$ will be merged as a D2S fused feature map $\mathcal{F}_{D2S} \in \mathbb{R} ^{C' \times H\times W}$:
\begin{equation}
    \mathcal{F}_{D2S}=\mathcal{F}_{e} \otimes \sigma\left[c(\mathcal{F}_{e}, \mathcal{F}_{i})\right] + \mathcal{F}_{e},
\end{equation}
where $\sigma(\cdot)$ and $c(\cdot)$ denote Sigmoid activate function and concatenation operations, respectively. 
Before the event feature is merged in the SPP module~\cite{zhao2017PSPNet}, standard Binary Cross Entropy (BCE) loss function and the ground-truth event data will be used for supervised learning. 
Furthermore, aiming to learn the whole model in an end-to-end fashion, the Cross Entropy (CE) loss from RGB branch will be merged with the BCE loss as:
\begin{equation}
\mathcal{L}=\mathcal{L}_{BCE}(V, \hat{V}) + \mathcal{L}_{CE}(Y, \hat{Y}),
\end{equation}
where $V$, $\hat{V}$, $Y$ and $\hat{Y}$ are the ground-truth and the predicted event volume, segmentation ground truth and prediction, respectively. 

\subsection{ISSAFE: Event-aware Domain Adaptation}

\begin{figure*}[t]
    \centering
    \vskip -1ex
    \includegraphics[width=0.8\textwidth]{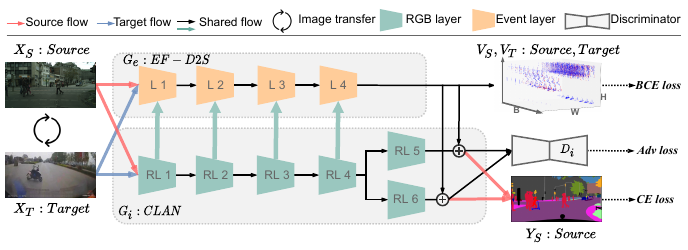}
    \caption{\small Architecture of the ISSAFE-EDA model with event-aware fusion in D2S mode.}
    \label{fig:event_fuse_CLAN_D2S}
\vskip -4ex
\end{figure*}

The source RGB images have labels and the target images do not, but the event-based data of both domains are available. Thus, we propose \textit{Event-aware Domain Adaptation (EDA)} to reduce the domain gap of normal and abnormal datasets by fusing the event-aware data, which can also explore the large-scale unlabeled accident images of our DADA-seg.
Compared to textured RGB images, the monochromatic event data, capturing only changes of intensity, is semantically more consistent in both domains, that denotes the homogeneous event features and thus can serve as a bridge to assist the RGB modal DA in the feature level. Based on this assumption, as shown in~\cref{fig:event_fuse_CLAN_D2S}, the ISSAFE-EDA consists of two branches, where the light-weight event-aware branch is the same as that in the aforementioned D2S fusion method and the RGB branch is constructed by the ResNet-101~\cite{he2016resnet} backbone referring to the CLAN~\cite{luo2019CLAN} model. When the source RGB image $X_S$ has ground-truth label $Y_S$, CE loss for the RGB branch $G_{i}$ in~\cref{fig:event_fuse_CLAN_D2S} is:
\begin{equation}
    \resizebox{.9\hsize}{!}{$\mathcal{L}_{CE}(G_{i}) 
    =-\sum_{h, w} \sum_{c \in C} Y_{S}^{(h, w, c)} \log \left(G_{i}(X_{S}^{(h, w, c)}) + G_{e}(X_{S}^{(h, w, c)}) \right)$}. 
\end{equation}
The source event-based data $E_S$ is learned from its ground truth $V_S$ via BCE in the event branch $G_{e}$:
\begin{equation}
    \mathcal{L}_{BCE}(G_e) 
    =-\sum_{h, w} \sum_{b \in B} V_{S}^{(h, w, b)} \log \left(G_{e}(E_{S}^{(h, w, b)})\right). 
\end{equation} 
The adversary loss between the RGB generator $G_i$ and discriminator $D_i$ according to the target image $X_T$ is:
\begin{equation}
\begin{aligned}
    \mathcal{L}_{adv}(G_i, D_i)=&-E[\log (D_i(G_i(X_{S})+G_e(X_{S})))] \\
    &-E[\log (1-D_i(G_i(X_{T})))].
\end{aligned}
\end{equation}
The final loss $\mathcal{L}_{EDA}$ of our ISSAFE-EDA model is combined from above 3 loss functions. Then, its training objective is:
\begin{equation}
    G_i^{*}, G_e^{*}, D_i^{*}=\arg \min_{G_i, G_e} \max _{D_i} \mathcal{L}_{EDA}(G_i, G_e, D_i).
\end{equation}
In this paper, we are making an early attempt to perform the cross-modal EDA from normal to abnormal scenes between two heterogeneous modalities. 

\section{Experiments}

\subsection{Datasets}\label{exp:dataset}

\begin{table}[!b]
\vskip -4ex
\caption{\small Statistics of datasets for experiments. The Merge3 dataset is combined with the Cityscapes, KITTI-360 and BDD3K, since these three datasets have the identical label mapping.}
\label{tab:stat_datasets}
\resizebox{\columnwidth}{!}{
\begin{tabular}{l|r|r|r|r|r|r}
\toprule
\textbf{Datasets} & \textbf{Cityscapes} & \textbf{KITTI-360} & \textbf{BDD3K} & \textbf{ApolloScape} & \textbf{Merge3} & \textbf{DADA-seg} \\ \midrule
\#training & 2,975 & 5,504 & 3,086 & 6,056 & 11,565 & - \\ 
\#evaluaion & 500 & 612 & 343 & 673 & 14,555 & 313 \\ \bottomrule
\end{tabular}
}

\end{table}

Since the synthesis of event data requires image pairs ($I_{t-1}$, $I_{t}$), two requirements to select a source dataset are: \textit{(i)} the anchor image $I_{t}$ has semantic annotation; and \textit{(ii)} the previous image $I_{t-1}$ is available. Finally, Cityscapes~\cite{cordts2016cityscapes}, KITTI-360~\cite{xie2016semantic}, BDD~\cite{yu2020bdd100k} and ApolloScape~\cite{wang2019apolloscape} are selected for our experiments. The statistics of the datasets are described in~\cref{tab:stat_datasets}.
The ApolloScape and KITTI-360 datasets have semantic annotations for each frame of their video sequences. Therefore, we only sample one anchor image every 10 frames from the video sequence to prevent overfitting cases. As only partial anchor images have annotations in the BDD dataset, we filtered it out based on the aforementioned two conditions, and termed it as \textit{BDD3K} in our work. Since the category definition of ApolloScape is different, so those models trained on it perform segmentation with only 16 overlapping categories~\cite{wang2019apolloscape}. 
As the target domain, our proposed DADA-seg dataset has 313 evaluation images from abnormal driving scenes, and the other unlabeled data were used to perform our EDA. All of these data including synthesized event-based data will be open-sourced to foster future research on accident scene understanding.

\subsection{Experimental Settings}
For efficiency reasons, we choose ResNet-18~\cite{he2016resnet} as the backbone and the main architecture from SwiftNet~\cite{orsic2019swiftnet}, which is also selected as the baseline model in this work. These models are trained with the Adam optimizer with a Learning Rate (LR) initialized to 4e-4 and dynamically adjusted by the cosine annealing LR scheduling strategy. The minimum LR of the last epoch is fixed in 1e-6. The weight decay of the LR is set to 1e-4. We use ImageNet pre-trained ResNet-18 to initialize the RGB encoder, and use the Kaiming initialization to initialize the whole Event branch as well as the decoder. For parameters of the pre-trained RGB encoder, we update them with a 4$\times$ smaller LR than the initialized parameters of the event encoder and the decoder, and apply a 4$\times$ smaller weight decay. The data augmentation operation includes a random scaling factor between 0.5 and 2, random horizontal flipping and random cropping with an output resolution of 1024$\times$512. We trained the model for 200 epochs with a batch size of 4 per GPU.

\begin{table}[!t]
\caption{\small Performance gap of models, which are trained and validated on Cityscapes and then evaluated on DADA-seg, both with 1024$\times$512 resolution.}
\label{tab:domain_gap}
\centering
\resizebox{\columnwidth}{!}{
\begin{tabular}{@{}llrrc@{}}
\toprule
\textbf{Network} & \textbf{Backbone} & \textbf{Cityscapes} & \textbf{DADA-seg} & \textbf{mIoU Gap} \\ \midrule \midrule
PSPNet~\cite{zhao2017PSPNet}            & MobileNetV2  & 70.2  & 17.1  & -52.5 \\
ERFNet~\cite{romera2017erfnet}          & ResNet-18    & 72.1  & 9.0   & -63.1 \\ 
SwiftNet~\cite{orsic2019swiftnet}       & ResNet-18    & 75.4  & 20.5  & -54,9 \\
DeepLabV3+~\cite{chen2018deeplabv3plus} & MobileNetV2  & 75.2  & 16.5  & -58.7 \\ 
DeepLabV3+~\cite{chen2018deeplabv3plus} & ResNet-50    & 79.0  & 19.0  & -60.0  \\ 
DeepLabV3+~\cite{chen2018deeplabv3plus} & ResNet-101   & 79.4  & 23.6  & -55.8  \\
DNL~\cite{yin2020dnl}                   & ResNet-50    & 79.3  & 15.7  & -63.6 \\
DNL~\cite{yin2020dnl}                   & ResNet-101   & 80.4  & 19.7  & -60.7 \\
OCRNet~\cite{yuan2019OCRNet}            & HRNetV2p-W18small & 77.1 & 20.5  & -56.6 \\
OCRNet~\cite{yuan2019OCRNet}            & HRNetV2p-W18 & 77.7  & 23.8  & -53.9  \\ 
OCRNet~\cite{yuan2019OCRNet}            & HRNetV2p-W48 & 80.6  & 24.9  & -55.7  \\ 
\bottomrule
\end{tabular}}
\vskip -3ex
\end{table}

\begin{figure*}[!t]
    \centering
    \includegraphics[width=0.99\textwidth]{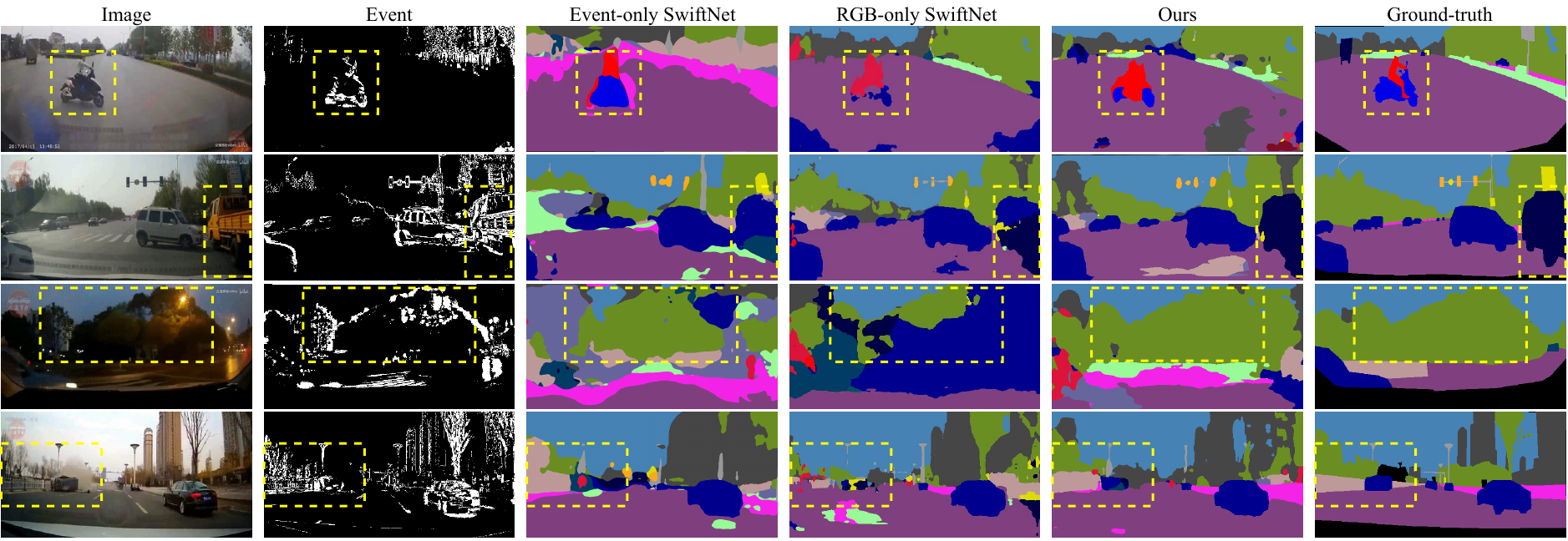}
    \vskip -1ex
    \caption{\small Contrastive examples of EF. The event data are presented as gray-scale frames. From top to bottom are accident scenes in different situations: motorist collision, car-truck collision, car collision at night time, and initial accident with an overturned car.}
    \label{fig:comparison_swiftnet}
\vskip -3ex
\end{figure*}

\subsection{Performance Gap}\label{exp:performance_gap}
To quantitatively evaluate the robustness of semantic segmentation algorithms, accuracy- and efficiency-oriented models are tested on the target dataset, as shown in \cref{tab:domain_gap}. For a fair comparison, when applicable, the results and model weights are provided by the respective publications. Overall, the large gaps show that SSA is still a challenging task for these top-performance models. As expected, although both large~\cite{chen2018deeplabv3plus, yuan2019OCRNet, yin2020dnl} and light-weight~\cite{zhao2017PSPNet, romera2017erfnet, orsic2019swiftnet} models gain high accuracy in the source domain, they heavily depend on the consistency between the training and the testing data, which are all normal scenes. It thus hinders their generalization ability and leads to a large performance degradation once taken to the abnormal scenes. 
Nonetheless, this comparison indicates that higher performance in the source domain still benefits performance in the target domain in most cases. In subsequent subsections, we perform ablation studies to verify the effectiveness of our proposed methods for reducing the large gap and improving the robustness in accident scenarios.

\begin{table}[t]
\caption{\small Comparison of different event representations and event fusion approaches. All models use ResNet-18 as backbone and are tested with 1024$\times$512 resolution.} 
\label{tab:event_fuse}
\centering
\resizebox{\columnwidth}{!}{
\begin{tabular}{@{}llllrr@{}}
\toprule
\textbf{Network} & \textbf{Input} & \textbf{Fusion} & \textbf{Event data} & \textbf{Cityscapes} & \textbf{DADA-seg} \\
\midrule \midrule
SwiftNet~\cite{orsic2019swiftnet}    & Event     & - & $B=1$         & 35.6   & 2.3    \\
SwiftNet~\cite{orsic2019swiftnet}    & Event     & - & $B=2$         & 36.0   & 19.7   \\
SwiftNet~\cite{orsic2019swiftnet}    & Event     & - & $B=18$        & 36.6   & 19.8   \\
\midrule
SwiftNet~\cite{orsic2019swiftnet}    & RGB       & - & -             & 69.2   & 20.1   \\
\midrule
\textbf{ISSAFE-S2D}      & RGB+Event & S2D & $B=1$   & 68.3   & 16.7   \\
\textbf{ISSAFE-S2D}      & RGB+Event & S2D & $B=2$   & 68.4   & 23.0   \\
\textbf{ISSAFE-S2D}      & RGB+Event & S2D & $B=18$  & 67.1   & 10.4   \\
\midrule
\textbf{ISSAFE-D2S}   & RGB+Event & D2S & $B=1$     & 69.0   & 24.5   \\
\textbf{ISSAFE-D2S}   & RGB+Event & D2S & $B=2$   & \textbf{69.4}   & \textbf{28.3}   \\ 
\textbf{ISSAFE-D2S}   & RGB+Event & D2S & $B=18$   & 68.8   & 24.5 \\
\bottomrule
\end{tabular}}
\vskip -4ex
\end{table}

\subsection{Ablation of EF}\label{exp:event_fusion}
\textbf{Quantitative Analysis.} As shown in~\cref{tab:event_fuse}, starting with event-only SwiftNet, where the event data are processed alone S2D without RGB image, the higher time bin $B$ brings better performance, and attains the mIoU of 36.6\% in the source domain and 19.8\% in the target domain. This indicates that the event data has certain interpretability for the segmentation of driving scenes. As a baseline, we train the SwiftNet with RGB only from scratch, which obtains 20.1\% mIoU in the target domain. Compared with it, our ISSAFE-S2D obtains an mIoU improvement of +2.9\%, while maintaining better performance in the source domain. When the event data is used as auxiliary information of the RGB branch, the model is improved in the moderate event representation (B=2), because others are too few or sparse for the RGB image. Likewise, we implement the ISSAFE-D2S, which brings over +8.2\% gain in the target domain when compared with the RGB-only baseline, meanwhile surpassing more than 10 state-of-the-art segmentation methods listed in~\cref{tab:domain_gap}.

\begin{table}[t]
\caption{\small Comparison between different source datasets and the mIoU gain on the target dataset.}
\label{tab:exp_dataset}
\centering
\begin{tabular}{@{}c|lccc@{}}
\toprule
\textbf{Dataset} & \textbf{Network} & \textbf{Source} & \textbf{DADA-seg} & \textbf{Gain} \\
\midrule \midrule
\multirow{2}{*}{Cityscapes~\cite{cordts2016cityscapes}} & SwiftNet~\cite{orsic2019swiftnet}    & 69.2  & 20.1 & \\
& \textbf{ISSAFE-D2S}   & 69.4 & 28.3 &\textbf{+8.2}  \\ \midrule
\multirow{2}{*}{BDD3K~\cite{yu2020bdd100k}} & SwiftNet~\cite{orsic2019swiftnet}    & 30.6  & 23.9 &\\
& \textbf{ISSAFE-D2S}   & 36.5 & 28.6  &\textbf{+4.7} \\ \midrule
\multirow{2}{*}{KITTI-360~\cite{xie2016semantic}} & SwiftNet~\cite{orsic2019swiftnet}    & 45.2  & 13.7 &\\
& \textbf{ISSAFE-D2S}   & 46.6 & 16.1 &\textbf{+2.4} \\ \midrule
\multirow{2}{*}{ApolloScape~\cite{wang2019apolloscape}} & SwiftNet~\cite{orsic2019swiftnet}    & 61.8  & 16.7 & \\
& \textbf{ISSAFE-D2S}   & 58.8 & 19.5 &\textbf{+2.8} \\ \midrule
\multirow{2}{*}{Merge3} & SwiftNet~\cite{orsic2019swiftnet}    & 50.3  & 28.5 & \\
& \textbf{ISSAFE-D2S}   & 61.4& \textbf{32.4}  &\textbf{+3.9} \\
\bottomrule
\end{tabular}
\vskip -3ex
\end{table}

\begin{figure*}[t]
    \centering
    \includegraphics[width=0.99\textwidth]{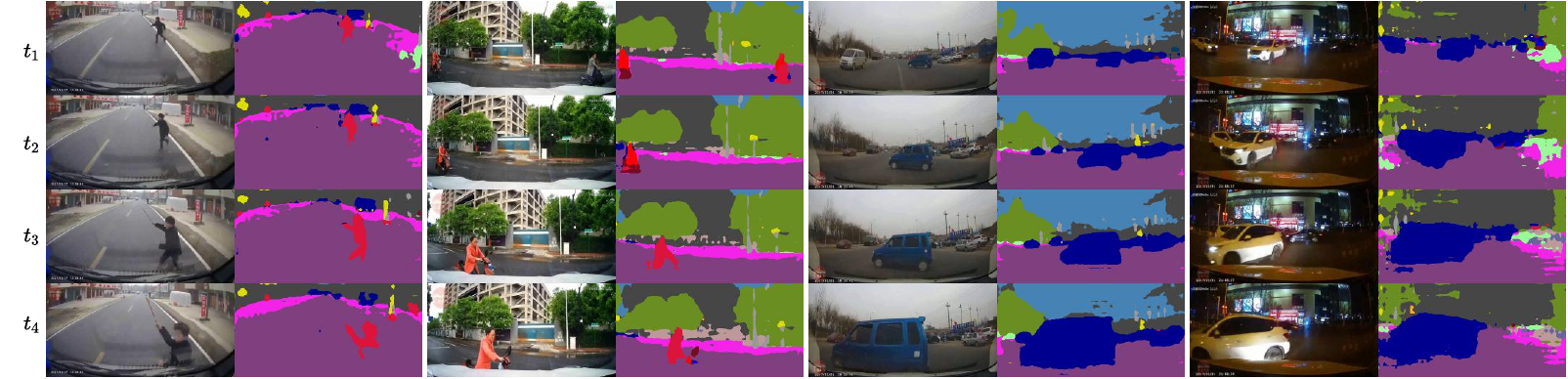}    
    \caption{\small Semantic segmentation results of our ISSAFE-EDA model on DADA-seg dataset. The columns correspond to the input images and output predictions of the sequence. 
    These traffic accidents occur under different natural conditions (in day- or night-time), and include various collided objects (pedestrians, cyclists, and cars).}
    \label{fig:results_CLAN}
\vskip -1ex
\end{figure*}

\begin{table*}[t]
\centering
\small \setlength{\tabcolsep}{2.5pt}
\caption{\small Performance comparison of DA strategies, where \textit{f} and \textit{i} represent the feature and image level.
The results of \textit{Source} and \textit{Target} are tested with 1024$\times$512 resolution, while \textit{Foreground classes} and \textit{Target$\dag$} are with 512$\times$256. To clearly showcase the effect of event-aware branch, the per-class IoU(\%) of ten foreground classes of \textit{Target} result are listed: \textit{Traffic Light, Traffic Sign, Pedestrian, Rider, Car, Truck, Bus, Train, Motorcycle} and \textit{Bicycle}. Note that the target dataset does not have any \textit{Train}.}
\label{tab:comparison_model}
\begin{tabular}{@{}ll|rrrrrrrrrr|ccc|ccc|ccc@{}}
\toprule
\multirow{2}{*}{\textbf{Network}} & \multirow{2}{*}{\textbf{Level}}  & \multicolumn{10}{c|}{\textbf{Foreground classes}} & \multicolumn{3}{c|}{\textbf{Target}$\dag$} & \multicolumn{3}{c|}{\textbf{Source}} & \multicolumn{3}{c}{\textbf{Target}} \\ \cmidrule{3-21}
& & TLi & TSi & Ped & Rid & Car & Tru & Bus & Tra & Mot & Bic & Acc & mIoU & fwIoU  & Acc & mIoU & fwIoU & Acc & mIoU & fwIoU   \\ \midrule \midrule
CLAN~\cite{luo2019CLAN}  & -   & 15.2 & 5.3 & 4.0 & 3.4 & 32.6 & 8.8 & 28.8 & - & 4.2 & 0.1 & 34.0 & 19.4 & 45.5 & 56.3 & 43.7 & 77.2 & 28.1 & 16.8 & 38.3 \\
CLAN~\cite{luo2019CLAN}  & $f$   & 17.2 & \textbf{21.5} & 8.4 & 6.3 & 63.5 & 33.4 & 33.1 & - & 3.7 & 6.2 & 46.3 & 31.7 & 67.2 & 70.4 & 62.4 & 87.0 & 40.1 & 28.8 & 63.8 \\
CLAN~\cite{luo2019CLAN}  & $f+i$ & 17.0 & 20.0 & 9.4 & 5.2 & 64.3 & 36.8 & 35.9 & - & 5.6 & \textbf{7.7} & 47.3 & 32.4 & 66.3 & 73.2 & \textbf{64.8} & 87.3 & 39.4 & 28.2 & 60.6 \\
\midrule
ISSAFE-OF & $f+i$ & \textbf{18.1} & 17.7 & 9.5 & 8.1 & 64.3 & 34.8 & 34.9 & - & 5.1 & 7.3 & \textbf{48.3} & \textbf{33.4} & \textbf{69.6} & 71.6 & 62.9 & 87.4 & 40.9 & 29.2 & 64.3 \\ \midrule
\textbf{ISSAFE-EDA} & $f+i$ & 17.0 & 19.5 & \textbf{10.0} & \textbf{8.8} & \textbf{65.6} & \textbf{39.5} & \textbf{39.7} & - & \textbf{6.1} & 7.0 & 48.2 & 33.1 & 68.2 & \textbf{73.2} & 63.9 & \textbf{87.5} & \textbf{42.1}  & \textbf{30.0}  & \textbf{64.5}  \\
\bottomrule
\end{tabular}
\vskip -3ex
\end{table*}

\textbf{On Diverse Datasets.}
For extensive verification of fusing event-based data, we conduct comparisons on four different source datasets, where the contrastive results are shown in~\cref{tab:exp_dataset}.
In general, our ISSAFE-D2S is capable of improving the segmentation robustness by integrating event-based data.
Based on the diversity of data, our ISSAFE-D2S trained from BDD3K gains $+4.7\%$ on the DADA-seg dataset, compared to the SwiftNet.
On the KITTI-360 and ApolloScape datasets, our model has also considerable improvements in the target domain, as $+2.4\%$ and $+2.8\%$, respectively.
In particular, the mIoU of our ISSAFE-D2S on the Merge3 dataset reaches the highest score with 32.4\%.
Overall, the results show that our proposal is consistently and significantly effective for enhancing the reliability of SSA.

\textbf{Qualitative Analysis.} As it is shown in~\cref{fig:comparison_swiftnet}, our ISSAFE-D2S model concentrates on the motion information, especially the foreground objects, such as the motorcycle and truck in the accident scenes. However, segmentation of night scenes is still challenging, although our method greatly benefits from event data, in contrast to the baseline. A case of the initial accident scene is presented as well. Our model can robustly segment the overturned car lying on the road after fence collision thanks to multi-modal cues.

Hence, the two data modalities are obviously complementary. When event cameras will not be triggered in static scenes, RGB cameras can perfectly capture the scene and provide sufficient textures. When RGB cameras puzzle over adverse scenes, \textit{i.e.} fast-moving objects or low-lighting environments, the event camera can provide auxiliary information for robustifying semantic segmentation. 

\subsection{Experiments of EDA}\label{Subsection4_2}
\textbf{Evaluation Metrics.} Our EDA is performed on two different levels, \textit{i.e.}, feature and/or image level. For a comprehensive quantitative analysis, we have adopted three different metrics~\cite{long2015FCN}, namely pixel accuracy (Acc), mean intersection over union (mIoU) and frequency weighted intersection over union (fwIoU), as shown in~\cref{tab:comparison_model}. 

\textbf{Quantitative Analysis.} Initially, the CLAN model adapted from virtual to real domain was tested directly on the DADA-seg dataset without any adjustments, also named source-only CLAN. 
Note that here a smaller resolution input can obtain higher accuracy in the target domain. There are two main reasons: images of DADA-seg are originally with low resolution, and a smaller resolution can obtain a larger receptive field with wider context understanding, which indicates that correct classification is more critical in accident scenes than delineating the boundaries. Afterwards, we train the CLAN model from scratch in Cityscapes and DADA-seg datasets to verify the feature- and feature-image-level DA, whereas the latter obtained the highest mIoU of 64.8\% in the source domain. To distinguish and eliminate the impact of diverse DA strategies, we utilize the CycleGAN~\cite{zhu2017cycleGAN} model to translate style of images from Cityscapes to DADA-seg and perform image-level DA between the two domains, which is termed as $i$ in~\cref{tab:comparison_model}.

As a result, our ISSAFE-EDA model obtains the highest performance in all three metrics on the DADA-seg dataset, and achieves the top accuracy of 30.0\% in mIoU, 42.1\% in Acc, and 64.5\% in fwIoU at the higher resolution. 
In order to understand the impact of event-aware motion feature, we list the per-class IoU results of all 10 foreground classes in \cref{tab:comparison_model}. Those results demonstrate that the foreground classes can indeed benefit more from event data. Besides, compared to the CLAN model without using event data, the improvement of our cross-modal ISSAFE-EDA model is consistent with our assumptions that the monochromatic event data can serve as bridge to adapt textured images of two domains towards robust semantic segmentation. 

\textbf{Comparison with Optical Flow.} We replace the event-based data with Optical Flow~(OF). For a fair comparison based on the same sparsity of data, we only utilize the traditional Farneback~\cite{farneback2003DOF} method to generate optical flow data. Our ISSAFE-OF model also obtains accuracy improvements, which further confirms our assumption regarding the effectiveness of motion features as complementary information for segmenting RGB images. However, our ISSAFE-EDA model can achieve better performance on the foreground classes.
Although both data are synthesized, motion features with higher time resolution can still be extracted from event data to boost foreground segmentation. Besides, event cameras have a high dynamic range to enhance perception in low-light conditions, which better conforms with our ISSAFE subject for improving road safety.

\textbf{Qualitative Analysis.} Some semantic segmentation results generated by our ISSAFE-EDA model are presented in \cref{fig:results_CLAN}. These traffic accidents occur under different natural conditions (in day or night time), and include various collided objects (pedestrians, cyclists, or cars). All these qualitative studies help to throw insightful hints on how to obtain reliable perception in accident scenes for IV systems.

\section{Conclusions}
In the paper, we present a new task and its relevant evaluation dataset with pixel-wise annotations, which serves as a benchmark to assess the robustness and applicability of semantic segmentation algorithms. The main objective is to improve the segmentation performance of complex scenes in the application of intelligent vehicles, and ultimately reduce traffic accidents and ensure the safety of all traffic participants.
As an initial solution, we have constructed the multi-modal segmentation model based on our \textit{ISSAFE} architecture by fusing event-based data in different modes. Our experiments show that event data can provide complementary information under normal and extreme driving situations to enhance RGB images, such as fine-grained motion information and low-light sensitivity. Even though our experiments are somewhat limited by the use of synthetic events due to the lack of corresponding event data in common annotated datasets, we have observed consistent and large accuracy gains for models learned on multiple datasets including Cityscapes, KITTI-360, BDD and ApolloScape.

Eventually, the SSA task is highly complicated and full of challenges that the current segmentation performance still has large development space. The unlabeled data in the DADA-seg dataset may be explored in future work through other learning paradigms, such as self-supervised and contrastive learning, so that we can gain more insights from the accident scenarios.

\bibliographystyle{IEEEtran}
\bibliography{main}
\end{document}